\pdfoutput=1

\documentclass[11pt]{article}

\usepackage[preprint]{acl}

\usepackage{times}
\usepackage{latexsym}

\usepackage[T1]{fontenc}

\usepackage[utf8]{inputenc}

\usepackage{microtype}

\usepackage{inconsolata}

\usepackage{graphicx}

\usepackage{multirow}
\usepackage{float}
\usepackage{enumitem}
\usepackage{amsmath}
\usepackage{amssymb}
\usepackage[linesnumbered,ruled,vlined]{algorithm2e}
\usepackage{array}
\usepackage{graphicx} 

\newtheorem{definition}{Definition}
\newtheorem{example}{Example}

\title{SimGRAG: Leveraging Similar Subgraphs for Knowledge Graphs \\Driven Retrieval-Augmented Generation}

\author{Yuzheng Cai\thanks{Equal contribution.}, Zhenyue Guo$^*$, Yiwen Pei, Wanrui Bian, Weiguo Zheng \\
        Fudan University\\
\texttt{\{yuzhengcai21, zhenyueguo23, ywpei23, wrbian23\}@m.fudan.edu.cn},\\
\texttt{zhengweiguo@fudan.edu.cn}
}

\begin{document}
\maketitle
\begin{abstract}
Recent advancements in large language models (LLMs) have shown impressive versatility across various tasks. 
To eliminate their hallucinations, retrieval-augmented generation (RAG) has emerged as a powerful approach, leveraging external knowledge sources like knowledge graphs (KGs). 
In this paper, we study the task of KG-driven RAG and propose a novel \textit{\underline{Sim}ilar \underline{G}raph Enhanced \underline{R}etrieval-\underline{A}ugmented \underline{G}eneration}  (SimGRAG) method.
It effectively addresses the challenge of aligning query texts and KG structures through a two-stage process: (1) query-to-pattern, which uses an LLM to transform queries into a desired graph pattern, and (2) pattern-to-subgraph, which quantifies the alignment between the pattern and candidate subgraphs using a graph semantic distance (GSD) metric.
We also develop an optimized retrieval algorithm that efficiently identifies the top-$k$ subgraphs within 1-second on a 10-million-scale KG.  
Extensive experiments show that SimGRAG outperforms state-of-the-art KG-driven RAG methods in both question answering and fact verification.
Our code is available at \url{https://github.com/YZ-Cai/SimGRAG}.
\end{abstract}

\section{Introduction}
\label{sec: introduction}

Pre-trained large language models (LLMs) are popular for diverse applications due to their generality and flexibility \cite{LLM-survey-2023, LLM-survey-2024, agent-survey-2024}. 
To avoid the hallucinations or outdated knowledge of LLMs \cite{LLM-hallucination, KAPING}, Retrieval-Augmented Generation (RAG) \cite{RAG-survey-2024-Feb, RAG-survey-2024-Jan} integrates LLMs with external knowledge sources to produce grounded outputs, where knowledge graphs (KGs) \cite{KG-survey-2022} have emerged as a valuable option \cite{graphRAG-survey}. 

\begin{figure}[t]
    \centering
    \includegraphics[width=\linewidth]{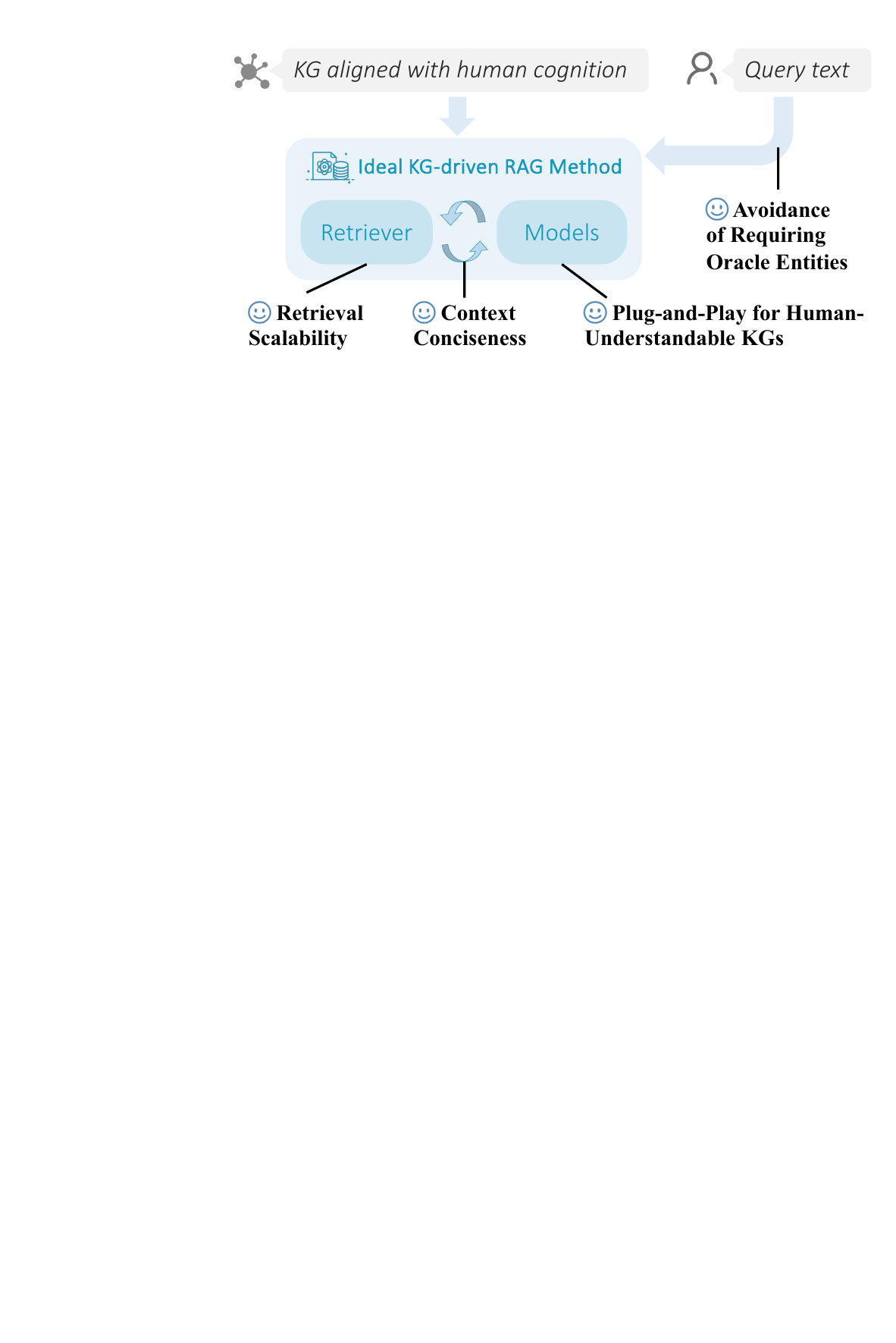}
    \caption{Ideal features for KG-driven RAG methods.}
    \label{fig: ideal features}
\end{figure}

For many KG-driven tasks, their KG schemas align with human cognition and can be read by humans.
In other words, a non-specialist can describe the knowledge using an intuitive graph structure.
In this paper, we follow existing KG-driven RAG methods \cite{KAPING, kggpt, KELP} and focus on such human-understandable KGs to enable the mimicking of human reasoning.
As shown in Figure~\ref{fig: ideal features}, an ideal approach should address the following features.

\begin{figure*}[t]
    \centering
    \vspace{-1mm}
    \includegraphics[width=\linewidth]{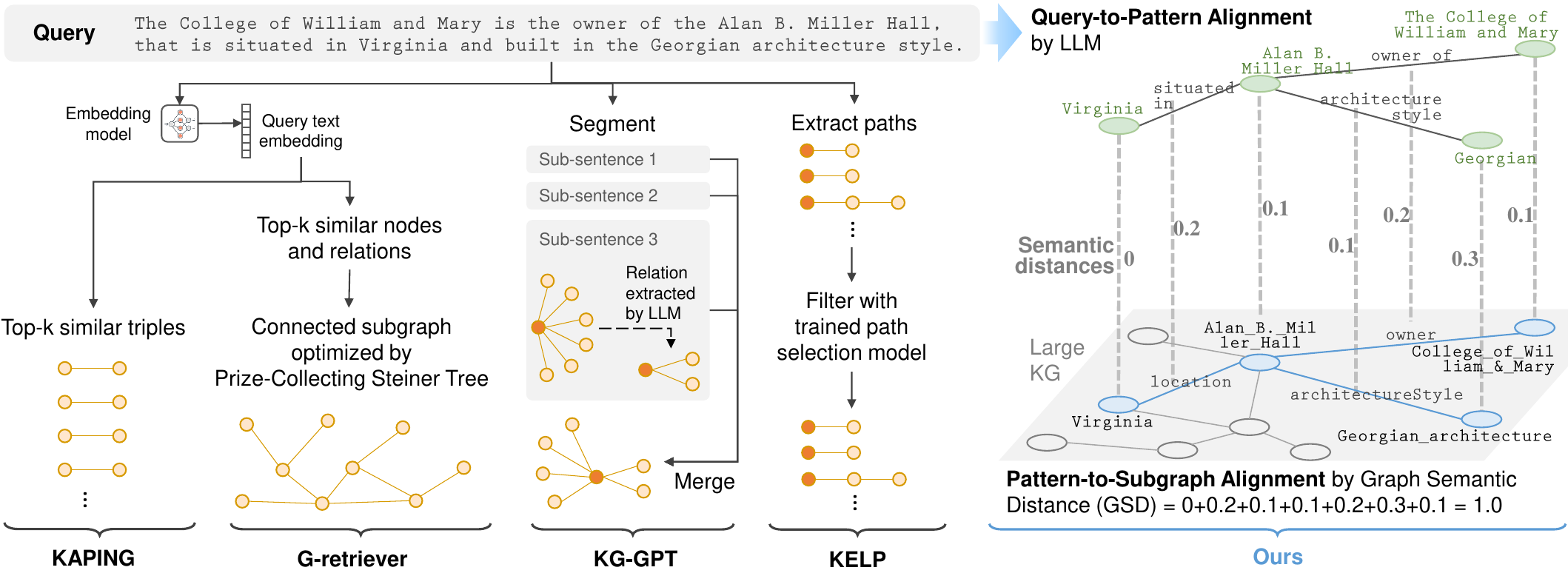}
    \caption{Comparison of mechanisms for aligning query text with KG structures. The example task is fact verification, where the query comes from FactKG dataset \cite{FactKG} with DBpedia \cite{DBpedia}.}
    \label{fig: comparison}
\end{figure*}

\textit{\underline{Plug-and-Play on Human-Understandable KGs.}}
To fully leverage the inherent generalization power of LLMs, an ideal approach should be easily deployable without additional training or fine-tuning for KGs that align with human cognition and can be interpreted by LLMs.
Otherwise, training a smaller and task-specific model on such KGs would be a more cost-effective alternative. 

\textit{\underline{Avoidance of Requiring Oracle Entities.}}
In real applications, users might not always know the precise entity IDs related to their queries. Thus, it would be better if a method naturally does not require users to specify the oracle entities.

\textit{\underline{Context Conciseness.}}
The retrieved subgraphs should focus on the most relevant and essential nodes and edges, ensuring clear contexts for LLMs. 

\textit{\underline{Retrieval Scalability.}}
An ideal algorithm should scale to large KGs with tens of millions of nodes and edges while maintaining acceptable latency.

Existing approaches typically follow a paradigm of retrieving subgraphs from the KG and feeding them into LLMs to generate the final response.
The critical challenge lies in effectively aligning query texts with the structural knowledge encoded in KGs.
Figure~\ref{fig: comparison} summarizes different mechanisms of existing approaches.
Specifically, 
\textit{(i)} KAPING \cite{KAPING} employs query text to directly retrieve isolated triples using their semantic embedding similarity, which struggles with multi-hop queries as the query embedding captures excessive information.
\textit{(ii)} G-retriever \cite{G-retriever} uses query text embeddings to retrieve similar entities and relations, then extracts a connected components in KG, which potentially cannot guarantee the best \textit{conciseness} of the retrieved subgraphs.
\textit{(iii)} KG-GPT \cite{kggpt} segments the query into sub-sentences but depends on the LLM to decide relations in KG that can match each sub-sentence, compromising \textit{scalability} as the number of candidate relations increases.
\textit{(iv)} KELP \cite{KELP} trains a path selection model to identify paths that align with the query text, lacking the \textit{plug-and-play} usability even on human-understandable KGs.

\begin{figure}[t]
    \centering
    \includegraphics[width=\linewidth]{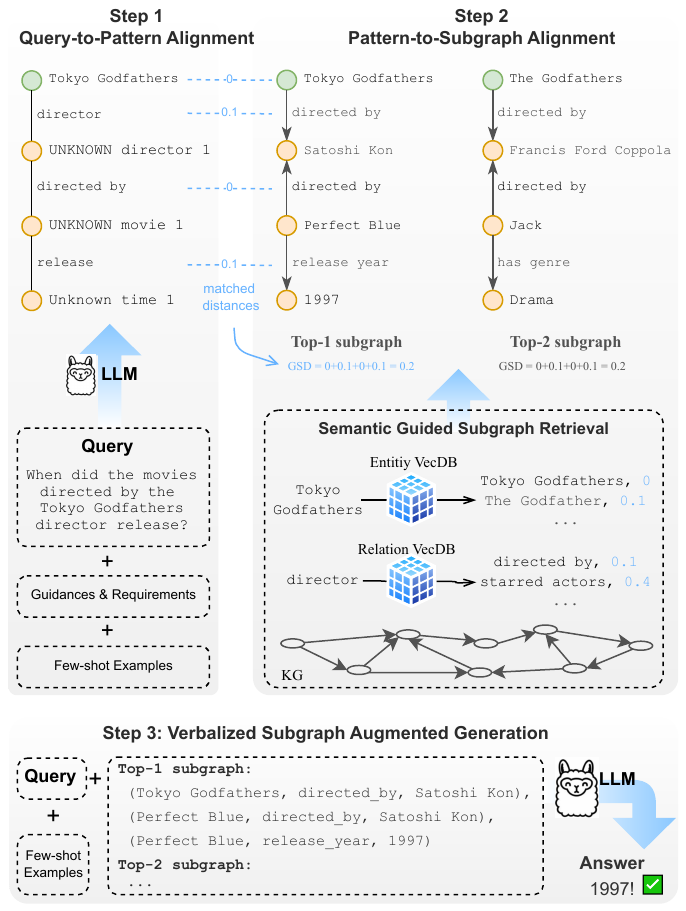}
    \caption{Overview of the SimGRAG method.}
    \label{fig: overview}
\end{figure}

In this paper, we introduce a novel approach, \textit{\underline{Sim}ilar \underline{G}raph Enhanced \underline{R}etrieval-\underline{A}ugmented \underline{G}eneration} (SimGRAG) method, for aligning query text with KG structures.
Figure~\ref{fig: overview} presents the overview with 3 steps.
\textit{(1) Query-to-Pattern Alignment.} 
We utilize an LLM to generate a pattern graph that aligns with the query text.
\textit{(2) Pattern-to-Subgraph Alignment.}
To retrieve the best subgraphs from KG that semantically align with the generated pattern graph, we introduce a novel metric termed \textit{Graph Semantic Distance} (GSD).
It quantifies the alignment by summing the semantic distances between corresponding nodes and relations in the pattern graph and the candidate isomorphic subgraphs.
For example, in Figure~\ref{fig: comparison}, the LLM generates a star-shaped pattern graph aligning with the query.
And the highlighted subgraph with the smallest GSD is considered as the best-aligned subgraph in KG.
\textit{(3) Verbalized Subgraph Augmented Generation.} 
Finally, the query and the retrieved subgraphs are passed to an LLM to generate the answer.

Different from KG-GPT \cite{kggpt} that leverages LLMs to filter relations within large KG, we only ask LLMs to generate a small pattern graph.
Also, our method targets subgraphs structurally and semantically aligned with the pattern, fundamentally differing from KAPING \cite{KAPING} and G-retriever \cite{G-retriever} that do not explicitly constrain subgraph structure or size. 
Our method can support more complex pattern graph structures, diverging from KELP \cite{KELP} that trains a path selection model limited to 1-hop or 2-hop paths.
Moreover, to retrieve the top-$k$ similar subgraphs w.r.t. the pattern graph with the smallest GSD, we further develop an optimized algorithm with an average retrieval time of less than one second per query on a 10-million-scale KG.

Our contributions are summarised as follows.
\begin{itemize}[leftmargin=*]
    \item We propose the query-to-pattern and pattern-to-subgraph alignment paradigm, ensuring the plug-and-play usability on human-understandable KGs and the context conciseness for LLMs.
    \item We define the graph semantic distance and develop an optimized subgraph retrieval algorithm to avoid requiring oracle entities and ensure retrieval scalability on million-scale KGs.
    \item Extensive experiments across different KG-driven RAG tasks confirm that SimGRAG outperforms state-of-the-art baselines.
\end{itemize}

\section{Related Work}
\label{sec: related work}

\paragraph{Knowledge Graph Meets Large Language Models.}
Recently, the pre-trained large language models have shown the ability to understand and handle knowledge graph (KG) related tasks \cite{KG-LLM-survey-2023-Aug, KG-LLM-survey-2023-Dec, KG-LLM-survey-2024-Jan, KG-LLM-survey-2024-TKDE, KG-LLM-survey-2024-IJCAI}, such as KG construction \cite{KG-construction-1}, KG completion \cite{KG-completion-1, KG-completion-2}, KG embedding \cite{KG-embedding-1}, and so on.
Furthermore, existing studies \cite{GNN-LLM-2024-Jan, GNN-LLM-2024-Feb, GNN-LLM-2024-April, GNN-LLM-2024-June} have tried to integrate LLMs with Graph Neural Networks (GNNs) to enhance modeling capabilities for graph data.

\paragraph{Retrieval-Augmented Generation.}
In practice, LLMs may produce unsatisfactory outputs due to their hallucination or inner outdated knowledge \cite{KAPING}.
Retrieval-Augmented Generation (RAG) \cite{RAG-survey-2024-Jan, RAG-survey-2024-Feb} is a promising solution that retrieves related information from external databases to assist LLMs.
Driven by documents, naive RAG approaches divide them into text chunks, which are embedded into dense vectors for retrieval.
There are a bunch of studies and strategies optimizing each step of the RAG process \cite{RAG-survey-2024-Feb}, including chunk division \cite{RAG-survey-2024-Jan}, chunk embedding \cite{RAG-embedding-1, RAG-embedding-2}, query rewriting \cite{RAG-rewriting-1}, document reranking \cite{RAG-survey-2024-Jan}, and LLM fine-tuning \cite{RAG-fintuning-1}.

\paragraph{Graph Retrieval-Augmented Generation.}
Graph Retrieval-Augmented Generation (GraphRAG) integrates graphs into RAG pipelines, which can be categorized into 10 domains, including knowledge graph (KG), document graph and so on \cite{GraphRAG-survey-1}.
GraphRAG methods may use existing graphs or construct graphs from other data source, such as building a knowledge graph (KG) from documents \cite{Distill-SynthKG}.
We focus on the KG-driven RAG scenario, which utilizes existing manually constructed KGs that used for retrieval in the RAG pipeline, as detailed as follows.

\paragraph{Knowledge Graph Driven Retrieval-Augmented Generation.}
The intricate structures of knowledge graphs (KGs) present significant challenges to traditional RAG pipelines, prompting the development of various techniques for graph-based indexing, retrieval, and generation \cite{graphRAG-survey}. 
As depicted in Figure~\ref{fig: comparison}, KAPING \cite{KAPING} retrieves KG triples most relevant to the query directly. 
KG-GPT \cite{kggpt} segments the query and presents LLMs with all candidate relations in the KG for decision-making. 
KELP \cite{KELP} trains a model to encode paths in the KG for selecting relevant paths, although it struggles to scale to structures more complex than 2-hop paths. 
G-Retriever \cite{G-retriever} first retrieves similar entities and relations, then constructs a connected subgraph optimized via the prize-collecting Steiner tree algorithm, and employs a GNN to encode the subgraph for prompt tuning with the LLM. 

\section{Preliminaries} 
\label{sec: preliminary}

A knowledge graph (KG) $\mathcal{G}$ is defined as a set of triples, i.e., $\mathcal{G} = \{(h, r, t) \mid h, t \in \mathcal{V}, r \in \mathcal{R} \}$, where $\mathcal{V}$ represents the set of entity nodes and $\mathcal{R}$ denotes the set of relations.
Given a knowledge graph $\mathcal{G}$ and a user query $\mathcal{Q}$, the task of \textit{Knowledge Graph Driven Retrieval-Augmented Generation} is to generate an answer $\mathcal{A}$ by leveraging both large language models and the retrieved evidence from $\mathcal{G}$. 
This task is general and encompasses a variety of applications, including but not limited to Knowledge Graph Question Answering (KGQA) and Fact Verification \cite{kggpt, KELP}.

An embedding model (EM) transforms a textual input $x$ to an $n$-dimensional embedding vector $z$ that captures its semantic meaning, i.e., $z = \text{EM}(x) \in \mathbb{R}^n$.
And the L2 distance between two vectors $z_1$ and $z_2$ is denoted by $\| z_1 - z_2 \|_2 \in \mathbb{R}$.

\section{The SimGRAG Approach}
\label{sec: approach}

Effectively aligning query text with the KG structures is a critical challenge.
In this section, we introduce a novel strategy that decomposes this alignment task into two distinct phases: query-to-pattern alignment and pattern-to-graph alignment.

\subsection{Query-to-Pattern Alignment}
\label{sec: query-to-pattern}

Given a query text $\mathcal{Q}$, we prompt the LLM to generate a pattern graph $\mathcal{P}$ consisting of a set of triples $\{(h_1, r_1, t_1), (h_2, r_2, t_2), \dots \}$ that align with the query semantics. 
We expect the LLM to interpret the user query thoughtfully, but we do not expect it to produce the exact same entities or relations appeared in the KG.

To guide the LLM in generating the desired patterns, our prompt first asks for the segmented phrases for each triple before generating all the triples.
As shown in Table~\ref{tab: query-to-pattern FactKG prompts}, it also includes a few explicit requirements.
To facilitate in-context few-shot learning \cite{few-shot-in-context}, we further manually construct a few examples (typically 12-shots) based on the characteristics of each KG, guiding the LLM to generate desired patterns. 

Such query-to-pattern alignment leverages the inherent understanding and instruction-following capabilities of LLMs.
Based on our experiments detailed in Section~\ref{sec: experiment}, the accuracy of the alignment can be defined as the proportion of queries that conform to the expected pattern under manual verification.
For queries involving up to 3 hops in the MetaQA \cite{MetaQA} and FactKG  \cite{FactKG} datasets, Llama 3 70B \cite{llama3} achieves the accuracies of 98\% and 93\%, respectively.
Thus, on KGs following human cognition which can be understood by humans, such alignment could be effectively performed by the LLM without the need for additional training, ensuring plug-and-play usability. 
But for certain KGs with specialized structures, it may be inevitable to further fine-tune the LLMs for mimicing domain-specific specialists, as discussed in Section~\ref{sec: limitations}.

\subsection{Pattern-to-Subgraph Alignment}
\label{sec: pattern-to-subgraph}

Given the generated pattern graph $\mathcal{P}$, our objective is to assess the overall similarity between $\mathcal{P}$ and a subgraph $\mathcal{S}$ in the knowledge graph $\mathcal{G}$.
Since the pattern $\mathcal{P}$ defines the expected structure of a subgraph, we leverage graph isomorphism to enforce structural constraints on the desired subgraph.

\begin{definition}[Graph Isomorphism]
The pattern graph $\mathcal{P}$ has a node set $V_\mathcal{P}$, while the subgraph $\mathcal{S}$ has a node set $V_\mathcal{S}$.
We say that $\mathcal{P}$ and $\mathcal{S}$ are isomorphic if there exists a bijective mapping $f: V_\mathcal{P} \rightarrow V_\mathcal{S}$ s.t. an edge $\langle u, v \rangle$ exists in $\mathcal{P}$ if and only if the edge $\langle f(u), f(v) \rangle$ exists in $\mathcal{S}$.
\end{definition}

Figure~\ref{fig: comparison} presents an isomorphism example.
Note that when checking graph isomorphism, we do not consider the edge direction, as different KGs may vary for the same relations. 
For instance, some KGs may express a relation such as ``person A directs movie B'', while others may use the reversed direction, ``movie B is directed by person A''.

After aligning the subgraph structure through graph isomorphism, we proceed to consider the semantic information of the nodes and relations. 
Similar to traditional text-driven RAG pipelines,
for each entity node $v$ and relation $r$ in both the pattern graph $\mathcal{P}$ and the subgraph $\mathcal{S}$, we obtain the corresponding embedding vectors $z$ as follows:
\begin{equation}
    z_v = \text{EM}(v), \quad z_r = \text{EM}(r)
\end{equation}

In this paper, we use the Nomic embedding model \cite{nomic}, which generates 768-dim semantic embeddings for nodes and relations.

For a subgraph $\mathcal{S}$ isomorphic to $\mathcal{P}$, the nodes and edges in $\mathcal{S}$ have a one-to-one mapping with those in $\mathcal{P}$. 
By computing the L2 distance between their embeddings, we use the pairwise matching distance \cite{pairwise-distance} to derive the following overall graph semantic distance.

\begin{definition}[Graph Semantic Distance, GSD]
\label{def: subgraph similarity}
Given the isomorphic mapping $f: V_\mathcal{P} \rightarrow V_\mathcal{S}$ between the pattern graph $\mathcal{P}$ and the KG subgraph $\mathcal{S}$, Graph Semantic Distance (GSD) is defined as follows, where $r_{\langle u,v \rangle}$ denotes the relation of the edge $\langle u,v \rangle$.
\end{definition}
{\fontsize{9.8}{12}\selectfont
\begin{align} 
    GSD(\mathcal{P}, \mathcal{S})=&\sum_{\text{node } v \in \mathcal{P}} \|z_v - z_{f(v)}\|_2 \\
    &+ \sum_{\text{edge } \langle u, v \rangle \in \mathcal{P}} \left\|z_{r_{\langle u,v \rangle}} - z_{r_{\langle f(u), f(v) \rangle}} \right\|_2, \notag
\end{align}}

\begin{example}
As illustrated in Figure~\ref{fig: comparison}, the highlighted subgraph in KG is isomorphic to the pattern graph. 
By computing the text similarity (i.e., embedding distance) between the matched nodes and edges, the resulting GSD is 1.0.
\end{example}

Focusing exclusively on isomorphic subgraphs guarantees conciseness.
Section~\ref{sec: retrieval} will provides algorithms to efficiently retrieve the top-$k$ isomorphic subgraphs with the smallest GSD in KG.

\begin{figure}[t]
    \centering
    \includegraphics[width=\linewidth]{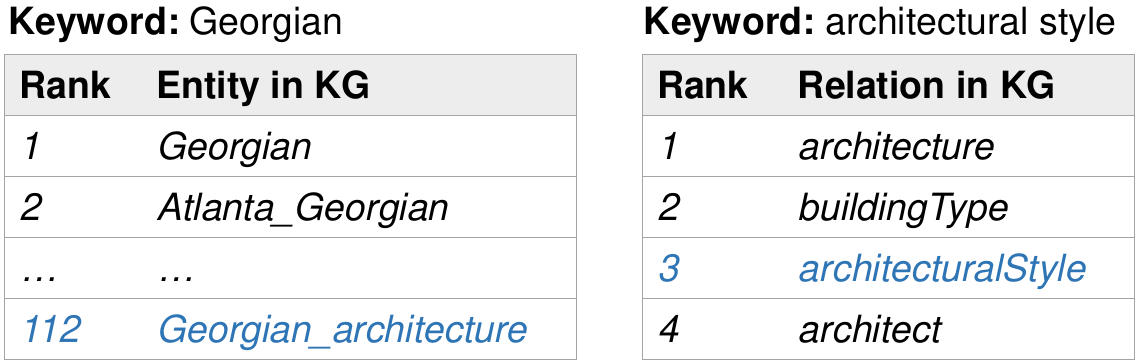}
    \caption{Semantic L2 distance rankings of a given keyword with entities (relations) in DBpedia \cite{DBpedia}, computed using the embeddings generated by the Nomic model \cite{nomic}.}
    \label{fig: semantic similarity}
\end{figure}

Furthermore, the joint use of graph isomorphism and semantic similarity effectively reduces noise.
In practice, KGs are often noisy, and even semantically similar entities or relations may not always constitute suitable evidence.
Figure~\ref{fig: semantic similarity} presents the distance rankings over the 10-million-scale DBpedia for the pattern graph in Figure~\ref{fig: comparison}.
There are numerous entities related to ``Georgian'', but only the entity ranked 112 contributes to the final subgraph. 
Similarly, for the relation ``architecture style'', only the relation ranked 3 is useful.
The proposed GSD metric can effectively incorporate somewhat distant entities or relations that still contribute valuable evidence to the overall subgraph, thereby eliminating the need for oracle entities.

\subsection{Generalization to Unknown Entities or Relations} \label{sec: unknown entity or relation}
In practice, some queries like ``Who is the director of the movie Her?'' may involve unknown entities. 
To address this, we extend the query-to-pattern alignment process by allowing the LLM to represent unknown entities or relations with unique identifiers such as ``UNKNOWN director 1'', as illustrated by the pattern graph $\mathcal{P}$ in Figure~\ref{fig: overview}.

In such cases, we further generalize the Graph Semantic Distance (GSD). 
Specifically, since the unknown entities or relations are ambiguous and difficult to match with corresponding entities or relations in the KG, we exclude them from the GSD computation. 
Given the isomorphic mapping $f: V_\mathcal{P} \rightarrow V_\mathcal{S}$ between the pattern graph $\mathcal{P}$ and the KG subgraph $\mathcal{S}$, we generalize GSD to:

{\fontsize{9}{14}\selectfont
\begin{align}
    GSD(\mathcal{P}, \mathcal{S})=&\sum_{\substack{\text{node } v \in \mathcal{P} \\ \text{s.t.} v \text{ is known}}} \|z_v - z_{f(v)}\|_2 \\
    &+ \sum_{\substack{\text{edge } \langle u, v \rangle \in \mathcal{P} \\ r_{\langle u, v \rangle} \text{ is known}}} 
    \|z_{r_{\langle u, v \rangle}} - z_{r_{\langle f(u), f(v) \rangle}}\|_2 \notag
\end{align}
}

\begin{example} 
As illustrated in Figure~\ref{fig: overview}, the top-1 subgraph from the KG yields a GSD of $0.2$.
\end{example}

\subsection{Verbalized Subgraph-Augmented Generation} 
\label{sec: generation}

Given the top-$k$ subgraphs with the smallest Graph Semantic Distance (GSD) from the KG, we now expect the LLM to generate answers to the original query based on these evidences. 
To achieve this, we append each retrieved subgraph $\mathcal{S}$ to the query text in the prompt. 
Each subgraph is verbalized as a set of triples $\{(h_1, r_1, t_1), (h_2, r_2, t_2), \dots\}$, as illustrated in Figure~\ref{fig: overview}. 
Additionally, to facilitate in-context learning, we also manually curate a few example queries (typically 12-shots) with their corresponding subgraphs and expected answers in the prompt.
Please refer to Appendix~\ref{app: prompts} for details.

\section{Semantic Guided Subgraph Retrieval}
\label{sec: retrieval}

Performing a brute-force search over all candidate subgraphs and computing the Graph Semantic Distance (GSD) for each one is computationally prohibitive. 
To address this, we propose a practical retrieval algorithm in Section~\ref{sec: naive retrieval}, which is further optimized for efficiency in Section~\ref{sec: optimized retrieval}.

\subsection{Top-$k$ Retrieval Algorithm}
\label{sec: naive retrieval}

Recent subgraph isomorphism algorithms often follow a \textit{filtering-ordering-enumerating} paradigm \cite{subgraph-matching-survey-2012, subgraph-matching-survey-2020, subgraph-matching-survey-2024}. 
To narrow down the potential search space, we first apply semantic embeddings to filter out unlikely candidate nodes and relations.
For each node $v_\mathcal{P}$ in the pattern graph $\mathcal{P}$, we retrieve the top-$k^{(n)}$ most similar entities from the knowledge graph $\mathcal{G}$, forming a candidate node set $C^{(n)}[v_\mathcal{P}]$. 
Similarly, for each relation $r_\mathcal{P}$, we extract the top-$k^{(r)}$ similar relations to form the candidate relation set $C^{(r)}[r_\mathcal{P}]$. 
Figure~\ref{fig: overview} illustrates an example of the candidate nodes and relations for the pattern graph node ``Tokyo Godfathers'' and the relation ``director''.
For unknown nodes or relations, as discussed in Section~\ref{sec: unknown entity or relation}, we treat all nodes or relations in $\mathcal{G}$ as candidates with a semantic distance of $0$.

The retrieval process is described in Algorithm~\ref{algo: naive}. 
Initially, lines~\ref{line: order start}-\ref{line: order end} organize all edges in $\mathcal{P}$ according to a DFS traversal order. 
For each candidate node $v_\mathcal{G}$ in the set $C^{(n)}[v^*_\mathcal{P}]$, we start an isomorphic mapping in lines~\ref{line: fix the first begin}-\ref{line: fix the first end} and iteratively expand the mapping using the \texttt{Expand} function until a valid mapping is found.
In function \texttt{Expand}, when matching the $i^{th}$ triple $(h_\mathcal{P}, r_\mathcal{P}, t_\mathcal{P})$ in the ordered triple list $L$, the node $h_\mathcal{P}$ is mapped to the corresponding node $h_\mathcal{G}$ in $\mathcal{G}$ via the partial mapping $f$. 
Then, lines~\ref{line: neighbor begin}-\ref{line: neighbor end} check each neighboring relation $r_\mathcal{G}$ and node $t_\mathcal{G}$ for $h_\mathcal{G}$ to see if they are valid candidates and do not contradict the existing mapping $f$.

\begin{algorithm}[t]
\caption{Top-$k$ Retrieval Algorithm}
\label{algo: naive}
\small
\KwIn{Pattern graph $\mathcal{P}$, knowledge graph $\mathcal{G}$, node candidates $C^{(n)}$, relation candidates $C^{(r)}$, and the parameter $k$.}
\KwOut{The top-$k$ subgraphs from $\mathcal{G}$ with the smallest GSD.}

Select start node $v^*_\mathcal{P}$ in $\mathcal{P}$ with the fewest candidates\;\label{line: order start}
$L \gets$ all triples of $\mathcal{P}$ in DFS traversal order from $v^*_\mathcal{P}$\;\label{line: order end}
$res \gets$ a priority queue maintaining the top-$k$ subgraphs with the smallest GSD\;

\ForEach{$v_\mathcal{G} \in C^{(n)}[v^*_\mathcal{P}]$ \label{line: fix the first begin}}{
    Expand($1, \{v^*_\mathcal{P}:v_{\mathcal{G}}\}$)\;\label{line: fix the first end}
}
\Return $res$\;

\vspace{2mm}
\SetKwFunction{FMain}{Expand}
\SetKwProg{Fn}{Function}{:}{}
\Fn{\FMain{$i, f$}}{
    \If{$f$ is a valid isomorphism mapping for $\mathcal{P}$}{
        Push the mapped subgraph $\mathcal{S}$ to $res$\;
        \Return\;
    }
    $(h_\mathcal{P}, r_\mathcal{P}, t_\mathcal{P}) \gets $ the $i^{th}$ triple in $L$ \label{line: ith triple}\;
    $h_\mathcal{G} \gets f(h_\mathcal{P})$\;
    \ForEach{$(r_G, t_G)$ s.t. $(h_G, r_G, t_G) \in \mathcal{G}$\label{line: neighbor begin}}{
        \If{$r_\mathcal{G} \in C^{(r)}[r_\mathcal{P}] \land t_\mathcal{G} \in C^{(n)}[t_\mathcal{P}]$}{
            \If{no contradiction for $t_\mathcal{P}$ in $f$}{
                Expand($i + 1, f \cup \{t_\mathcal{P}: t_\mathcal{G}\}$)\;\label{line: neighbor end}
            }
        }
    }
}
\end{algorithm}

\subsection{Optimized Retrieval Algorithm}
\label{sec: optimized retrieval}

Despite the filtering approach, the above algorithm still suffers from a large search space, especially when there are too many candidate nodes and relations.
As we only need the top-$k$ subgraphs with the smallest GSD, we propose an optimized strategy that can prune unnecessary search branches.

Assume that during the expansion of the $i^{\text{th}}$ edge in $L$, the partial mapping from $\mathcal{P}$ to the knowledge graph $\mathcal{G}$ is represented by $f$. 
Suppose there exists an isomorphic mapping $f'$ that can be completed by future expansion, resulting in a subgraph $\mathcal{S}$ with $GSD(\mathcal{P}, \mathcal{S})$.
It can be decomposed into four terms, where $L[1:i]$ denotes the first $i-1$ triples in $L$ and $L[i:]$ denotes the remaining triples.

{\fontsize{9}{13}\selectfont
\begin{align} 
&GSD(\mathcal{P}, \mathcal{S}) = \Delta^{(n)}_{\text{mapped}} + \Delta^{(n)}_{\text{remain}} + \Delta^{(r)}_{\text{mapped}} + \Delta^{(r)}_{\text{remain}}, \label{eq: total GSD}\\
&\Delta^{(n)}_{\text{mapped}}
= \sum_{\substack{\text{node } v_\mathcal{P} \in \mathcal{P} \\ \text{ mapped in } f}} \| z_{v_\mathcal{P}} - z_{f(v_\mathcal{P})} \|_2, \\ 
&\Delta^{(n)}_{\text{remain}} =  \sum_{\substack{\text{node } v_\mathcal{P} \in \mathcal{P} \\ \text{ not mapped in } f}} \| z_{v_\mathcal{P}} - z_{f'(v_\mathcal{P})} \|_2, \label{eq: future nodes}\\
&\Delta^{(r)}_{\text{mapped}} = \sum_{( h_\mathcal{P}, r_\mathcal{P}, t_\mathcal{P} ) \in L[1:i]} \| z_{r_\mathcal{P}} - z_{r_{\langle f(h_\mathcal{P}), f(t_\mathcal{P}) \rangle}} \|_2,\\
&\Delta^{(r)}_{\text{remain}} = \sum_{( h_\mathcal{P}, r_\mathcal{P}, t_\mathcal{P} ) \in L[i:]} \| z_{r_\mathcal{P}} - z_{r_{\langle f'(h_\mathcal{P}), f'(t_\mathcal{P}) \rangle}} \|_2. \label{eq: future edges} 
\end{align}
}

For Equations~(\ref{eq: future nodes}) and (\ref{eq: future edges}), notice that

{\fontsize{8.5}{11}\selectfont
\begin{align} 
\Delta^{(n)}_{\text{remain}} &\geq \sum_{\substack{\text{node } v_\mathcal{P} \in \mathcal{P} \\ \text{ not mapped in } f}} \min_{v_\mathcal{G} \in C^{(n)}[v_\mathcal{P}]} \| z_{v_\mathcal{P}} - z_{v_\mathcal{G}} \|_2 \triangleq X. \label{eq: future node bounds} \\
\Delta^{(r)}_{\text{remain}} &\geq \sum_{( h_\mathcal{P}, r_\mathcal{P}, t_\mathcal{P} ) \in L[i:]} \min_{r_\mathcal{G} \in C^{(r)}[r_\mathcal{P}]} \| z_{r_\mathcal{P}} - z_{r_\mathcal{G}} \|_2 \triangleq Y.
\label{eq: future edge bounds}
\end{align}
}

Combining Equations~(\ref{eq: total GSD}), (\ref{eq: future node bounds}), and (\ref{eq: future edge bounds}), we have

{\fontsize{9}{12}\selectfont
\begin{equation}
GSD(\mathcal{P}, \mathcal{S}) \geq \Delta^{(n)}_{\text{mapped}} + \Delta^{(r)}_{\text{mapped}} + X + Y \triangleq B. 
\end{equation}
}

When the lower bound $B$ exceeds the largest GSD of the top-$k$ subgraphs in current priority queue $res$, any subgraph $\mathcal{S}$ completed through future expansion will never become the desired top-$k$ subgraphs.
That is, the current partial mapping $f$ can be safely discarded, effectively pruning subsequent unnecessary search branches.

Moreover, to reduce the largest GSD in the top-$k$ priority queue $res$ for more pruning opportunities, we adopt a greedy strategy that prioritizes matching more promising subgraphs earlier.
Specifically, for lines~\ref{line: fix the first begin}-\ref{line: fix the first end}, we can process the nodes $v_\mathcal{G} \in C^{(n)}[v^*_\mathcal{P}]$ in ascending order of their distances. 
In line~\ref{line: neighbor begin} of the \texttt{Expand} function, the neighboring relation and node $(r_G, t_G)$ with the smaller sum of $\|z_{t_P} - z_{t_G}\|_2 + \|z_{r_P} - z_{r_G}\|_2$ will be expanded earlier.

By combining the pruned and greedy expansion strategies, the optimized algorithm is guaranteed to produce the same results as the top-$k$ retrieval algorithm without any loss in solution quality.
The experiments in Section~\ref{sec: efficacy} show that the optimized algorithm significantly accelerates retrieval.

\section{Experiments}
\label{sec: experiment}

We conduct experiments on tasks of \textit{Knowledge Graph Question Answering (KGQA)} and \textit{Fact Verification}.

\begin{table*}[t]
\centering
\small
\setlength{\tabcolsep}{4.5mm}
\begin{tabular}{l|c|c|c|c|c|c|c}
\hline
\multirow{2}{*}{\textbf{Method}} & \multicolumn{3}{c|}{\textbf{MetaQA} (Hits@1)} & \multicolumn{2}{c|}{\textbf{PQ} (Hits@1)} & \textbf{WC2014} & \textbf{FactKG} \\ \cline{2-6}
                & \textbf{1-hop} & \textbf{2-hop} & \textbf{3-hop} & \textbf{2-hop} & \textbf{3-hop} & (Hits@1) &  (Accuracy)   \\ \hline
\multicolumn{8}{c}{\textit{Supervised task-specific methods}} \\ \hline
EmbedKGQA        & 97.5                  & 98.8                  & 94.8      & - & - & -             & -              \\
NSM                    & 97.1                  & 99.9                  & 98.9      & - & - & -             & -              \\
UniKGQA           & 97.5                  & 99.0                  & 99.1        & - & - & -           & -              \\ 
Transfernet           & 97.5                  & 100                  & 100      & - & - & -             & -              \\ 
GEAR              & -                  & -                  & -             & - & - & -      & 77.7              \\ \hline
\multicolumn{8}{c}{\textit{Pre-trained LLMs}} \\ \hline
ChatGPT                 & 60.0                  & 23.0                  & 38.7    & - & - & -               & 68.5          \\
Llama 3 70B              & 56.7                  & 25.2                  & 42.3    & - & - & -               & 68.4          \\ \hline
\multicolumn{8}{c}{\textit{KG-driven RAG with training (Llama 3 70B)}} \\ \hline
KELP$^\dagger$                 & 94.7                  & 96.0                  & -  & - & - & -                    & 73.3          \\ 
G-Retriever$^\dagger$          & 98.5                  & 87.6                  & 54.9    & 61.8     & 46.7      & 67.5                  & 61.4        \\ \hline
\multicolumn{8}{c}{\textit{KG-driven RAG without training (Llama 3 70B)}} \\ \hline
KAPING                & 90.8                 & 71.2                 & 43.0     & 41.0     & 52.1      & 88.1            &   75.5            \\
KG-GPT$^\dagger$              & 93.6                 & 93.6                 & 88.2        & 86.1 & 42.5 & 71.1           & 69.5          \\
SimGRAG (ours)         & 98.0                 & 98.4                 & 97.8       & 88.7     & 78.6      & 98.1           & 86.8          \\ 
\hline
\end{tabular}
\caption{Performance comparison of different approaches, where $^\dagger$ denotes 
we provide oracle entities as it is the default setting of a method.
Each reported value serves as an upper bound for the result obtained without oracle entities.
Appendix~\ref{app: entity leak} presents more discussions.}
\label{tab: overall result}
\end{table*}

\subsection{Tasks and Datasets}
\label{sec: datasets}

\paragraph{Knowledge Graph Question Answering.}
We use the MoviE Text Audio QA dataset (MetaQA) \cite{MetaQA} related to the field of movies.
All the queries in the test set are adopted for evaluation, consisting of Vanilla 1-hop, 2-hop, and 3-hop question-answering in the same field.
We also use the PathQuestions dataset (PQ) \cite{PathQuestions} developed from Freebase \cite{Freebase} consisting of 2-hop and 3-hop queries, and the WorldCup2014 dataset (WC2014) \cite{WC2014} with sports-domain KGs.

\paragraph{Fact Verification.}
We adopt the FactKG dataset \cite{FactKG}, in which colloquial and written style claims
can be verified using the DBpedia \cite{DBpedia}.
All statements in the test set are used in the evaluation, and a method should return \textit{Supported} or \textit{Refuted} after verification.

Please refer to Appendix~\ref{app: datasets} for detailed statistics and examples of the tasks and datasets.

\subsection{Baselines}
\label{sec: baselines}

The included baselines are briefly introduced as follows.
Please refer to Appendix~\ref{app: implementations} for more details.

\paragraph{Supervised task-specific models.}
State-of-the-art models for KGQA include EmbedKGQA \cite{EmbedKGQA}, NSM \cite{NSM}, UniKGQA \cite{UniKGQA}, and Transfernet \cite{Transfernet}.
They are trained on the MetaQA training set and evaluated by the test accuracy.
For fact verification, the KG version of GEAR \cite{GEAR} is trained on the FactKG training set.

\paragraph{Pre-trained LLMs.}
For both tasks, we evaluate two popular LLMs, ChatGPT \cite{ChatGPT} and Llama 3 70B \cite{llama3}, using 12-shots without any provided evidence.

\paragraph{KG-driven RAG with training.}
Recent method KELP \cite{KELP} trains the retriever over the training set, while G-retriever \cite{G-retriever} trains a graph neural network (GNN) to integrate query texts and subgraph evidences.

\paragraph{KG-driven RAG without training.}
Both KAPING \cite{KAPING} and KG-GPT \cite{kggpt} only require retrieval subgraphs from the KGs without any training or fine-tuning.

\begin{table*}[t]
  \centering
  \small
  \setlength{\tabcolsep}{4.8mm}
  \begin{tabular}{c|ccc|cc|c|c}
    \hline
 \multirow{2}{*}{} & \multicolumn{3}{c|}{\textbf{MetaQA} (Hits@1)} & \multicolumn{2}{c|}{\textbf{PQ} (Hits@1)} & \textbf{WC2014} & \textbf{FactKG} \\ \cline{2-6}
       & \textbf{1-hop} & \textbf{2-hop} & \textbf{3-hop} & \textbf{2-hop} & \textbf{3-hop} &  (Hits@1) &  (Accuracy)   \\ \hline
   shot=4 & 98.6 & 96.5 & 92.8 & 90.9 & 78.3 & 97.2 & 84.0 \\
  shot=8 & 98.3 & 96.4 & 98.8 & 90.3 & 79.5 & 96.3 & 87.9 \\
  shot=12 & 98.0 & 98.4 & 97.8 & 88.7 & 78.6 & 98.1 & 86.8 \\\hline
    $k=1$ & 95.2 & 98.2 & 97.0 & 90.4 & 76.9 & 93.2 & 88.1 \\
    $k=2$ & 98.0 & 97.9 & 97.6 & 90.3 & 77.7 & 93.6 & 87.6 \\
    $k=3$ & 98.0 & 98.4 & 97.8 & 88.7 & 78.6 & 98.1 & 86.8 \\\hline
Llama3-70B &	98.0 &	98.4	& 97.8	& 88.7	& 78.6	& 98.1 & 86.8	\\
Phi4-14B &	92.7 &	99.5 &	90.8	&	92.2 &	83.4	& 91.6 & 86.1 \\
Qwen2.5-72B &	98.6 &	99.8 &	98.2	&	88.7 &	77.7	& 97.5  & 83.6\\\hline
  \end{tabular}
  \caption{Performance of the SimGRAG method by varying the number of few-shot examples, the parameter $k$ for semantic guided subgraph retrieval, and different LLMs.}
  \label{tab: vary k}
  \label{tab: vary shots}
  \label{tab: vary LLM}
\end{table*}

\subsection{Comparative Results}
\label{sec: comparison}

As summarized in Table~\ref{tab: overall result}, supervised task-specific methods outperform KG-driven RAG approaches that require additional training.
Notably, supervised task-specific methods generally require smaller model sizes and lower training costs, making them a more cost-effective option in practice.

Directly using LLMs leads to the poorest performance.
As for KG-driven RAG methods without additional training, SimGRAG shows substantially higher Hits@1 and accuracy in most cases. 
In fact, SimGRAG performs comparably to supervised task-specific models and even outperforms the supervised GEAR method on the FactKG dataset.

Moreover, the performance gap between SimGRAG and other RAG approaches becomes larger as the complexity of the questions increases on the MetaQA dataset. 
As discussed in Section~\ref{sec: pattern-to-subgraph}, the combined use of graph isomorphism and semantic similarity effectively reduces noise and ensures conciseness, thus benefiting the performance of SimGRAG for 2-hop and 3-hop questions.

\subsection{Ablation Studies}
\label{sec: ablation}

\paragraph{Few-shot in-context learning.}
Table~\ref{tab: vary shots} evaluates SimGRAG method by varying the number of examples in the prompts, used in both pattern-to-graph alignment and verbalized subgraph-augmented generation. 
For the simplest MetaQA 1-hop questions, performance is not sensitive to the number of shots. 
In contrast, for more complex queries like those in the MetaQA 3-hop, PQ 3-hop, and FactKG datasets, we observe significant improvements when increasing from 4 to 8 shots.

\paragraph{Parameter $k$ for semantic guided subgraph retrieval.}
Table~\ref{tab: vary k} reports the impact of parameter $k$ for retrieving top-$k$ subgraphs with the smallest graph semantic distance.
For MetaQA 1-hop questions, setting $k=1$ leads to a significant drop in Hits@1, since many movies share exactly the same title, and retrieving fewer subgraphs makes it more difficult to cover the ground-truth answer.
For MetaQA 2-hop and 3-hop questions, the choice of $k$ has a negligible impact on performance. 
Conversely, increasing $k$ leads to a slight decrease in accuracy on the FactKG dataset, since the top-1 subgraph is often sufficient and including more subgraphs will introduce noise for LLM.

\paragraph{Choice of Large Language Models.}
We also evaluate the proposed SimGRAG using two additional open-source LLMs, including Phi4-14B \cite{phi4} and Qwen2.5-72B \cite{qwen25}. 
The results in Table~\ref{tab: vary LLM} demonstrate that SimGRAG is generally robust across different LLMs. Even using the Phi-4 14B model without any training or finetuning, SimGRAG remains competitive with existing methods. 
Also, SimGRAG offers a plug-and-play solution on human-understandable KGs across various LLMs, and we expect its performance to improve with future LLM advancements.

\paragraph{Query pattern structure.}
As outlined in Appendix~\ref{app: pattern structures}, we categorize query pattern structures into six classes and show the distributions in Table~\ref{tab: query percent}.
Table~\ref{tab: overall result} confirm that SimGRAG outperforms RAG baselines on multi-hop path queries, and it is also better on WC2014 dataset that contains 22\% 2-hop conjunction queries.
By further experiments on each category of queries for FactKG dataset, SimGRAG achieves the accuracies of 89\%, 88\%, and 85\% on 2-hop conjunction, 3-hop conjunction, and 3-hop star queries, respectively.

\begin{table}[t]
\small
\centering
\vspace{-2.5mm}
\setlength{\tabcolsep}{0.9mm}
\begin{tabular}{l|rrr|rr|r}
\hline
\multicolumn{1}{c|}{\multirow{2}{*}{\textbf{Dataset}}} & \multicolumn{3}{c|}{\textbf{Path}} & \multicolumn{2}{c|}{\textbf{Conjunction}} & \multicolumn{1}{c}{\textbf{Star}} \\
\cline{2-7}
\multicolumn{1}{c|}{} & 1-hop & 2-hop & 3-hop & 2-hop & 3-hop & 3-hop \\
\hline
MetaQA 1-hop & 100\% & 0 & 0 & 0 & 0 & 0 \\
\hline
MetaQA 2-hop & 0 & 100\% & 0 & 0 & 0 & 0 \\
\hline
MetaQA 3-hop & 0 & 0 & 100\% & 0 & 0 & 0 \\
\hline
PQ 2-hop & 0 & 100\% & 0 & 0 & 0 & 0 \\
\hline
PQ 3-hop & 0 & 0 & 100\% & 0 & 0 & 0 \\
\hline
WC2014 & 64\% & 14\% & 0 & 22\% & 0 & 0 \\
\hline
FactKG & 32\% & 28\% & 5\% & 17\% & 8\% & 10\% \\
\hline
\end{tabular}
\caption{Distribution of query pattern structures.}
\label{tab: query percent}
\end{table}

\subsection{Error Analysis}
\label{sec: error analysis}

Table~\ref{tab: error analysis} summarizes the error distribution across the three main steps of the SimGRAG method. 
For detailed error examples, please refer to Appendix~\ref{app: error analysis}.

Many errors occur during the query-to-pattern alignment step, where the LLM fails to follow the given instructions and examples, thereby generating the undesired pattern graphs. 
Generally, both 2-hop and 3-hop queries roughly have consistent proportion of errors.
But there are more errors on 1-hop queries, since we use the same few-shot examples for all MetaQA queries, which are all about 2-hop or 3-hop patterns. They make the LLM sometimes generate 2-hop patterns for 1-hop queries.

As the complexity of the queries increases in the MetaQA dataset, we also observe a higher incidence of errors in the subgraph-augmented generation step, since it is more difficult for the LLM to accurately extract relevant information for a complex question from the retrieved subgraphs.

On the FactKG dataset, errors are also encountered during the pattern-to-subgraph alignment. 
In these cases, while the LLM generates reasonable subgraphs in line with the guidance, mismatches occur because the ground-truth subgraphs have different structures and thus cannot be successfully aligned, as illustrated in Appendix~\ref{app: error analysis}.

\begin{table}[t]
\centering
\small
\setlength{\tabcolsep}{1.2mm}
\begin{tabular}{c|ccc|c}
\hline
   \multirow{2}{*}{\textbf{Step}} & \multicolumn{3}{c|}{\textbf{MetaQA}} & \multirow{2}{*}{\textbf{FactKG}} \\ \cline{2-4}
                & \textbf{1-hop} & \textbf{2-hop} & \textbf{3-hop} &  \\ \hline
Query-to-pattern &              89\%                                                   & 36\%                                                            & 31\%                                                            & 49\%            \\
Pattern-to-subgraph     & 0\%                                                           & 0\%                                                            & 0\%                                                            & 24\%            \\
Augmented generation         &                                       11\%                         & 64\%                                                            & 69\%                                                            & 27\%      \\
\hline
\end{tabular}
\caption{The statistics of errors from different steps.}
\label{tab: error analysis}
\end{table}

\begin{figure}[t] 
\centering 
\includegraphics[width=\linewidth]{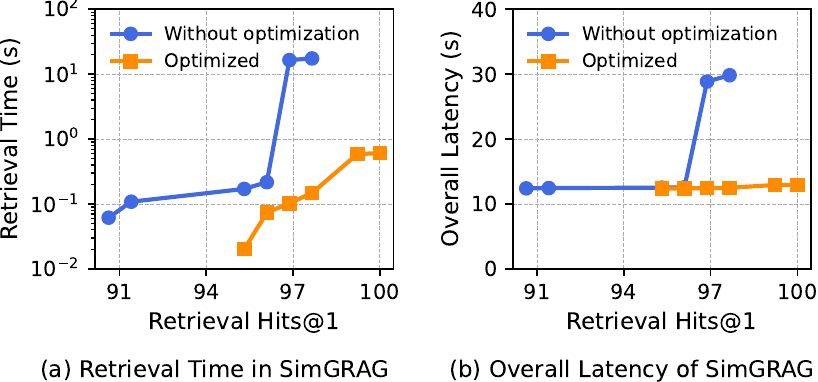} 
\caption{Pareto optimal curves for retrieval.} 
\label{fig: time cost} 
\end{figure}

\begin{table}[t]
  \centering
  \small
  \setlength{\tabcolsep}{1.6mm}
  \begin{tabular}{l|rrr|r}
    \hline
\multirow{2}{*}{} & \multicolumn{3}{c|}{\textbf{MetaQA}} & \multirow{2}{*}{\textbf{FactKG}} \\ \cline{2-4}
                & \textbf{1-hop} & \textbf{2-hop} & \textbf{3-hop} &  \\ \hline
    Vector search & 0.02 & 0.02 & 0.02 & 0.59 \\
    Optimized retrieval & 0.0006 & 0.0007 & 0.002 & 0.15 \\ \hline
    \textbf{Total} & 0.02 & 0.02 & 0.02 & 0.74 \\ \hline
  \end{tabular}
  \caption{Semantic guided subgraph retrieval time (s).}
  \label{tab: time cost}
\end{table}

\begin{table}[!htp]
    \centering
    \small
    \setlength{\tabcolsep}{2.5mm}
    \begin{tabular}{l|ccc|c}
        \hline
        \multirow{2}{*}{\textbf{Method}} & \multicolumn{3}{c|}{\textbf{MetaQA}} & \multirow{2}{*}{\textbf{FactKG}} \\
        \cline{2-4}
        & 1-hop & 2-hop & 3-hop & \\
        \hline
        KELP & 3.6 & 4.8 & - & 5.4 \\
        G-Retriever & 4.1 & 4.3 & 4.4 & 5.4 \\
        KAPING & 3.3 & 5.9 & 8.8 & 10.6 \\
        KG-GPT & 10.1 & 12.3 & 13.1 & 13.3 \\
        Ours (4-shot) & 5.5 & 5.9 & 7.3 & 10.2 \\
        Ours (12-shot) & 9.0 & 9.1 & 11.9 & 14.2\\
        \hline
    \end{tabular}
    \caption{Comparison of average query latency (s).}
    \label{tab: overall latency}
\end{table}

\subsection{Retrieval Efficiency} 
\label{sec: efficacy}

As discussed in Section~\ref{sec: retrieval}, we first perform a vector search to obtain the top-$k^{(n)}$ candidate nodes and top-$k^{(r)}$ candidate relations. 
Table~\ref{tab: time cost} reports the average retrieval time per query, in which the vector search time dominates the total time. 
On the 10-million-scale DBpedia KG from the FactKG dataset, the overall retrieval time is 0.74 seconds per query, highlighting the efficiency and scalability of the optimized retrieval algorithm.

Additionally, we conduct a grid search over the parameters $k^{(n)}$ and $k^{(r)}$ to compare the top-$k$ retrieval and the optimized algorithms. 
Please refer to Appendix~\ref{app: grid search} for detailed setups. 
Figure~\ref{fig: time cost}(a) presents the Pareto optimal curves, which plot the trade-off between average retrieval time and retrieval Hits@1.
The results clearly show that the optimized retrieval algorithm significantly improves the performance, particularly in scenarios where a higher retrieval Hits@1 is desired in practice. 
Also, Figure~\ref{fig: time cost}(b) shows the overall latency for the proposed SimGRAG method, in which the optimized algorithm guarantees reasonable latency. 

\subsection{Overall Latency}
\label{sec: overall latency}

We run each method on a NVIDIA A6000 GPU using Ollama 4-bit quantization for Llama3 70B.
Table~\ref{tab: overall latency} reports the average latency for answering each query.
Generally, our method has similar latency compared with others using the default 12-shot in-context learning. 
It could be much faster with 4-shot learning while still providing competitive performance, as confirmed by Table~\ref{tab: vary shots}.

\section{Conclusion}
\label{sec: conclusion}

In this paper, we investigate the problem of KG-driven RAG and introduce a novel SimGRAG approach that effectively aligns query texts with KG structures. 
For query-to-pattern alignment, we employ an LLM to generate a pattern graph that aligns with the query text.
For pattern-to-subgraph alignment, we introduce the Graph Semantic Distance (GSD) metric to quantify the alignment between the desired pattern and the underlying subgraphs in the KG.
Additionally, we propose an optimized algorithm to retrieve the top-$k$ similar subgraphs with the smallest GSD, improving retrieval efficiency and scalability. 
Extensive experiments demonstrate that SimGRAG consistently outperforms existing KG-driven RAG approaches.

\section*{Acknowledgments}

This work was substantially supported Key Projects of the National Natural Science Foundation of China (Grant No. U23A20496).
Weiguo Zheng is the corresponding author.

\section*{Limitations} 
\label{sec: limitations}

The performance of SimGRAG method is closely tied to the underlying capabilities of the large language model (LLM). 
Specifically, the method relies heavily on the ability of LLMs to understand and follow instructions effectively in both steps of the query-to-pattern alignment and verbalized subgraph-augmented generation. 
Thus, the performance of SimGRAG can be substantially degraded when utilizing lower-quality or less capable LLMs, especially in scenarios involving more complex queries that demand advanced reasoning skills. 

Furthermore, following the characteristics of KGs used by existing studies \cite{KAPING, kggpt, KELP, G-retriever}, we also assume that our input KG aligns with human cognition.
It is a key requirement for the plug-and-play usability for the SimGRAG method.
However, when using industrial domain-specific KGs which diverge significantly from commonly used schemas, it is challenging for LLMs to predict the desired nodes, edges or pattern structures during the query-to-pattern alignment stage.
It is still under exploration how effectively fine-tuning LLMs can help to align generated patterns with such special KG structures.
Also, we can include the specialized KG schema in prompts, guiding LLMs to generate patterns more likely isomorphic to desired subgraphs in the KG.

Additionally, for domain-specific KGs, linking query entities to corresponding candidate entities in the KG could be challenging, particularly when the embedding model has not been trained on such data. 
Therefore, rather than relying on a plug-and-play embedding model, future work may fine-tune the embedding models on domain-specific data or explore alternative entity linking approaches.

\bibliography{custom}

\begin{table*}[t]
    \centering
    \small
    \setlength{\tabcolsep}{2.2mm}
    \begin{tabular}{c|c|cccc}
    \hline
    \textbf{Dataset} & \textbf{Underlying KG} & \textbf{\# Entity nodes} & \textbf{\# Relation edges} & \textbf{\# Entity type} & \textbf{\# Relation type}                        \\
    \hline
    MetaQA     & MetaQA                     & 43,234                   & 269,482           & -                       & 9                             \\
    PathQuestions     & PathQuestions                    & 2,215                   & 3,321           & -                       & 14                             \\
    WC2014     & WC2014                     & 1,127                   & 6,482           & -                       & 6                             \\
    FactKG     & DBpedia                     & 9,912,183               & 42,879,918                & 467                     & 522  \\ \hline
    \end{tabular}
    \caption{Statistics of the knowledge graphs.}
    \label{tab: dataset}
\end{table*}

\begin{table*}[t]
\centering
\small
\setlength{\tabcolsep}{9mm}
\begin{tabular}{ccc}
\hline
\textbf{Head}                          & \textbf{Relation} & \textbf{Tail}    \\
\hline
Champagne for Caesar                   & has genre         & Comedy           \\
High Risk                              & starred actors    & Lindsay Wagner   \\
Married to It                          & directed by       & Arthur Hiller    \\
The Adventures of Huckleberry   Finn   & directed by       & Michael Curtiz   \\
The Amazing Spider-Man 2               & directed by       & Marc Webb        \\
The Eiger Sanction                     & starred actors    & Clint Eastwood   \\
The Exterminating Angel                & has tags          & luis buñuel      \\
The Life and Times of Hank   Greenberg & has genre         & Documentary      \\
The Slumber Party Massacre             & directed by       & Amy Holden Jones \\
Tokyo Godfathers                       & release year      & 2003   \\
\hline
\end{tabular}
\caption{Example triples in the knowledge graph of MetaQA dataset.}
\label{tab: MetaQA KG examples}
\end{table*}

\begin{table*}[!htp]
\centering
\small
\begin{tabular}{m{3.7cm}<{\centering}m{10cm}}
    \hline
        \textbf{Dataset} & \multicolumn{1}{c}{\textbf{Example questions}} \\
    \hline
        MetaQA 1-hop &  1. what films did Michelle Trachtenberg star in? \newline 2. what are some words that describe movie Lassie Come Home? \newline 3. who is the director of The Well-Digger's Daughter? \\
    \hline
        MetaQA 2-hop &  1. which movies have the same actor of Jack the Bear? \newline 2. which movies share the same director of I Wanna Hold Your Hand? \newline 3. what were the release dates of Eric Mandelbaum written films? \\
    \hline
        MetaQA 3-hop &  1. who wrote movies that share directors with the movie Unbeatable? \newline 2. what genres do the movies that share directors with Fish Story fall under? \newline 3. who acted in the films written by the screenwriter of The Man Who Laughs? \\
    \hline
\end{tabular}
\caption{Example questions in the MetaQA dataset.}
\label{tab: MetaQA query examples}
\end{table*}

\newpage
$\quad$

\newpage
\appendix

\section{Details of Tasks and Datasets}
\label{app: datasets}

Table~\ref{tab: dataset} summarizes the statistics of the underlying knowledge graph used for each dataset.

\subsection{Knowledge Graph Question Answering}

For the task of Knowledge Graph Question Answering, we use the MoviE Text Audio QA dataset (MetaQA) \cite{MetaQA}, PathQuestions dataset (PQ) \cite{PathQuestions}, and the WorldCup2014 dataset (WC2014) \cite{WC2014}.

MoviE Text Audio QA dataset (MetaQA) is designed for research on question-answering systems on knowledge graphs \cite{MetaQA}.
It provides a knowledge graph about movies, where entities include movie names, release years, directors, and so on, while the relations include starred actors, release year, written by, directed by, and so on.
The queries are composed of Vanilla 1-hop, 2-hop, and 3-hop question answering in the field of movies. 
For the test set of MetaQA dataset, there are 9,947 questions for 1-hop, 14,872 for 2-hop, and 14,274 for 3-hop.
Table~\ref{tab: MetaQA KG examples} shows some example triples in the knowledge graph provided in the MetaQA \cite{MetaQA} dataset, while Table~\ref{tab: MetaQA query examples} are some example questions in the dataset.
The MetaQA dataset is released under the Creative Commons Public License.

PathQuestions dataset (PQ) is built on Freebase KG \cite{Freebase}, which contains 1,908 2-hop path queries and 5,198 3-hop path queries \cite{PathQuestions}.
Table~\ref{tab: PQ KG examples} shows some example triples in the PathQuestions knowledge graph, while Table~\ref{tab: PQ query examples} are some example questions.
It is under a Creative Commons Attribution 4.0 International Licence.

WorldCup2014 dataset (WC2014) contains a knowledge graph about football players that participated in FIFA World Cup 2014 \cite{WC2014}. 
There are 10,162 queries of WC2014, which is a mixture of 6,482 single-relation questions, 1,472 two-hop path questions, and 2,208 conjunctive questions.
Table~\ref{tab: WC2014 KG examples} shows some example triples in the WC2014 knowledge graph, while Table~\ref{tab: WC2014 query examples} are some example questions.
It is under a Creative Commons Attribution 4.0 International Licence.

\begin{table*}[t]
\centering
\small
\setlength{\tabcolsep}{5.5mm}
\begin{tabular}{ccc}
\hline
\textbf{Head}                          & \textbf{Relation}      & \textbf{Tail}                          \\
\hline
eleanor\_of\_provence                    & children              & beatrice\_of\_england                    \\
manuel\_i\_of\_portugal                   & gender                & male                                   \\
joan\_crawford                          & spouse                & phillip\_terry                         \\
barbara\_of\_portugal                    & spouse                & ferdinand\_vi\_of\_spain                  \\
empress\_myeongseong                     & cause\_of\_death        & regicide                               \\
frederica\_of\_mecklenburg-strelitz       & spouse                & ernest\_augustus\_i\_of\_hanover            \\
henri\_victor\_regnault                  & gender                & male                                   \\
adelaide\_of\_lowenstein\_wertheim\_rosenberg & children              & maria\_josepha\_of\_portugal              \\
prince\_frederick\_duke\_of\_york\_and\_albany & place\_of\_death        & london                                 \\
mary\_boleyn                             & spouse                & william\_carey\_1490                     \\
\hline
\end{tabular}
\caption{Example triples in the knowledge graph of PQ dataset.}
\label{tab: PQ KG examples}
\end{table*}

\begin{table*}[t]
\centering
\small
\begin{tabular}{m{4cm}<{\centering}m{10cm}}
    \hline
        \textbf{Dataset} & \multicolumn{1}{c}{\textbf{Example questions}} \\
    \hline
        PQ 2-hop &  1. john\_b\_kelly\_sr's son's job? \newline 2. what is the sex of spouse of mary\_stuart\_countess\_of\_bute? \newline 3. where does virginia\_heinlein's spouse work for? \\
    \hline
        PQ 3-hop &  1. who is the mom of father of mary\_of\_teck's heir? \newline 2. what is the name of the gender of son of henry\_i\_of\_england's mother? \newline 3. ferdinand\_ii\_of\_aragon's parent's heir's nation? \\
    \hline
\end{tabular}
\caption{Example questions in the PQ dataset.}
\label{tab: PQ query examples}
\end{table*}

\begin{table*}[!htp]
\centering
\small
\setlength{\tabcolsep}{13.5mm}
\begin{tabular}{ccc}
\hline
\textbf{Head}                          & \textbf{Relation}      & \textbf{Tail}                          \\
\hline
Esseid\_BELKALEM                             & plays\_in\_club &   Watford\_FC                     \\
      Frank\_LAMPARD                           & plays\_for\_country &    England                      \\
JOAO\_MOUTINHO                          & plays\_in\_club        & AS\_Monaco                             \\
Agustin\_ORION                              & plays\_position &   Goalkeeper                       \\
Rickie\_LAMBERT                         & is\_aged               & 32                                     \\
             Pedro\_RODRIGUEZ               & plays\_position &  Forward                           \\
HENRIQUE                                & is\_aged               & 27                                     \\
OGC\_Nice                                  & is\_in\_country &  France                             \\
Andres\_GUARDADO                        & wears\_number          & 18                                     \\
 Michel\_VORM                          & plays\_position & Goalkeeper                              \\
\hline
\end{tabular}
\caption{Example triples in the knowledge graph of WC2014 dataset.}
\label{tab: WC2014 KG examples}
\end{table*}

\begin{table*}[!htp]
\centering
\small
\begin{tabular}{m{4cm}<{\centering}m{10cm}}
    \hline
        \textbf{Dataset} & \multicolumn{1}{c}{\textbf{Example questions}} \\
    \hline
        WC2014 1-hop &  1. which football club does Alan\_PULIDO play for? \newline 2. what position does Alan\_PULIDO play? \newline 3. which country is the soccer team Atletico\_Madrid based in? \\
    \hline
        WC2014 2-hop &  1. which professional foootball team has a player from Belgium? \newline 2. where is the football club that Rafael\_MARQUEZ plays for? \newline 3. which country does Mathieu\_DEBUCHY play professional in? \\
    \hline
        WC2014 Conjunction &  1. name a player who plays at Forward position at the club Tigres\_UANL? \newline 2. who are the Italy players at club US\_Citta\_di\_Palermo? \newline 3. which Portugal footballer plays at position Goalkeeper? \\
    \hline
\end{tabular}
\caption{Example questions in the WC2014 dataset.}
\label{tab: WC2014 query examples}
\end{table*}

\subsection{Fact Verification}
For the task of fact verification, we use the FactKG dataset \cite{FactKG} that contains 5 different types of fact verification: One-hop, Conjunction, Existence, Multi-hop, and Negation, while all of them can be verified using the DBpedia knowledge graph \cite{DBpedia}.
Its test set contains 9,041 statements to be verified.
Table~\ref{tab: FactKG KG examples} shows some example triples in the DBpedia, while Table~\ref{tab: FactKG query examples} are some example statements in the FactKG test set.
The FactKG dataset is licensed with CC BY-NC-SA 4.0.

\begin{table*}[t]
\centering
\small
\setlength{\tabcolsep}{2.4mm}
\begin{tabular}{ccc}
\hline
\textbf{Head}                                                & \textbf{Relation} & \textbf{Tail}                      \\
\hline
Berlin                                                       & country           & Germany                            \\
United States                                                & governmentType    & Republic                           \\
Harry Potter                                                 & author            & J. K. Rowling                        \\
Albert Einstein                                              & award             & Nobel Prize in Physics   \\
Terrance Shaw                                                & college           & Stephen F. Austin State University \\
Association for the Advancement of Artificial Intelligence & type              & Scientific   society               \\
Nvidia                                                       & industry          & Computer hardware         \\
\hline
\end{tabular}
\caption{Examples triples in the DBpedia knowledge graph used for FactKG dataset.}
\label{tab: FactKG KG examples}
\end{table*}

\begin{table*}[t]
\centering
\small
\setlength{\tabcolsep}{34mm}
\begin{tabular}{c}
\hline
\textbf{Example statements} \\
\hline
    1. It was Romano Prodi who was the prime minister. \\
    2. Are you familiar with Terrance Shaw? He also attended college. \\
    3. Yes, Anastasio J. Ortiz was the Vice President.\\
\hline
\end{tabular}
\caption{Example statements from the FactKG dataset.}
\label{tab: FactKG query examples}
\end{table*}

\begin{table*}[!htp]
\centering
\small
\begin{tabular}{|>{\arraybackslash}m{14.7cm}|}
\hline
You need to segment the given query then extract the potential knowledge graph structures.\\
\\
\textbf{Notes)}\\
1). Use the original description in the query with enough context, NEVER use unspecific words like 'in', 'appear in', 'for', 'of' etc.\\
2). For nodes or relations that are unknown, you can use the keyword 'UNKNOWN' with a unique ID, e.g., 'UNKNOWN artist 1', 'UNKNOWN relation 1'.\\
3). Return the segmented query and extracted graph structures strictly following the format:\\
\hspace{10mm}\{
        "divided": [
            "segment 1",
            ...
        ],
        "triples": [
            ("head", "relation", "tail"),
            ...
        ]
    \}\\
4). NEVER provide extra descriptions or explanations, such as something like 'Here is the extracted knowledge graph structure'.\\
\\
\textbf{Examples)}\\
1. query: "the actor in Flashpoint also appears in which films"\\
\hspace{3.5mm}output: \{\\
\hspace{14mm}"divided": [\\
\hspace{18mm}"the actor in Flashpoint",\\
\hspace{18mm}"this actor also appears in another films",\\
\hspace{14mm}],\\
\hspace{14mm}"triples": [\\
\hspace{18mm}("UNKNOWN actor 1", "actor of", "Flashpoint"),\\
\hspace{18mm}("UNKNOWN actor 1", "actor of", "UNKNOWN film 1"),\\
\hspace{14mm}]\\
\hspace{3.5mm}\}\\
2. query: ...\\
\hspace{3.5mm}output: ...\\
\\
\textbf{Your task)}\\
Please read and follow the above instructions and examples step by step\\
query: \{\{QUERY\}\}\\
\hline
\end{tabular}
\caption{The query-to-pattern alignment prompt used for KGQA task.}
\label{tab: query-to-pattern KGQA prompts}
\end{table*}

\section{Prompts}
\label{app: prompts}

For query-to-pattern alignment, Table~\ref{tab: query-to-pattern KGQA prompts} shows the prompt for KGQA tasks, including MetaQA, PathQuestions and WC2014 datasets.
Table~\ref{tab: query-to-pattern FactKG prompts} shows the prompt for the fact verification task, i.e., FactKG dataset.

For verbalized subgraph-augmented generation, Table~\ref{tab: answer KGQA prompts} shows the prompt for KGQA tasks, including MetaQA, PathQuestions and WC2014 datasets.
Table~\ref{tab: answer FactKG prompts} shows the prompt for the fact verification task, i.e., FactKG dataset.

At each step for processing each dataset, our prompts utilize exactly the same guidances and few-shot examples across different query pattern structures.
For example, all 1/2/3-hop queries in the MetaQA dataset share the identical prompt at the step of query-to-pattern alignment.

\begin{table*}[t]
\centering
\small
\begin{tabular}{|>{\arraybackslash}m{14.5cm}|}
\hline
You need to segment the given query then extract the potential knowledge graph structures.\\
\\
\textbf{Notes)}\\
1). Use the original description in the query with enough context, NEVER use unspecific words like 'in', 'appear in', 'for', 'of' etc.\\
2). For nodes or relations that are unknown, you can use the keyword 'UNKNOWN' with a unique ID, e.g., 'UNKNOWN artist 1', 'UNKNOWN relation 1'.\\
3). Return the segmented query and extracted graph structures strictly following the format:\\
\hspace{10mm}\{
        "divided": [
            "segment 1",
            ...
        ],
        "triples": [
            ("head", "relation", "tail"),
            ...
        ]
    \}\\
4). NEVER provide extra descriptions or explanations, such as something like 'Here is the extracted knowledge graph structure'.\\
\\
\textbf{Examples)}\\
1. query: "The College of William and Mary is the owner of the Alan B. Miller Hall, that is situated in Virginia."\\
\hspace{3.5mm}output: \{\\
\hspace{14mm}"divided": [\\
\hspace{18mm}"The College of William and Mary is the owner of the Alan B. Miller Hall",\\
\hspace{18mm}"Alan B. Miller Hall is situated Virginia",\\
\hspace{14mm}],\\
\hspace{14mm}"triples": [\\
\hspace{18mm}("The College of William and Mary", "owner", "Alan B. Miller Hall"),\\
\hspace{18mm}("Alan B. Miller Hall", "situated in", "Virginia"),\\
\hspace{14mm}]\\
\hspace{3.5mm}\}\\

2. query: ...\\
\hspace{3.5mm}output: ...\\
\\
\textbf{Your task)}\\
Please read and follow the above instructions and examples step by step\\
query: \{\{QUERY\}\}\\
\hline
\end{tabular}
\caption{The query-to-pattern alignment prompt used in FactKG dataset.}
\label{tab: query-to-pattern FactKG prompts}
\end{table*}

\begin{table*}[t]
\centering
\small
\begin{tabular}{|>{\arraybackslash}m{14.5cm}|}
\hline
Please answer the question based on the given evidences from a knowledge graph. \\
\\
\textbf{Notes)}\\
1). Use the original text in the valid evidences as answer output, NEVER rephrase or reformat them.\\
2). There may be different answers for different evidences. Return all possible answer for every evidence graph, except for those that are obviously not aligned with the query.\\
3). You should provide a brief reason with several words, then tell that the answer.\\
\\
\textbf{Examples)}\\
1. query: "who wrote films that share actors with the film Anastasia?"\\
\hspace{3.5mm}evidences: \{\\
\hspace{14mm}"graph [1]": [\\
\hspace{18mm}("Anastasia", "starred\_actors", "Ingrid Bergman"),\\
\hspace{18mm}("Spellbound", "starred\_actors", "Ingrid Bergman"),\\
\hspace{18mm}("Spellbound", "written\_by", "Ben Hecht"),\\
\hspace{14mm}],\\
\hspace{14mm}"graph [2]":[\\
\hspace{18mm}("Anastasia", "starred\_actors", "John Cusack"),\\
\hspace{18mm}("Floundering", "starred\_actors", "John Cusack"),\\
\hspace{18mm}("Floundering", "written\_by", "Peter McCarthy"),\\
\hspace{14mm}]\\
\hspace{3.5mm}\}\\
\hspace{3.5mm}answer: According to graphs [1][2], the writter is Ben Hecht or Peter McCarthy.

2. query: ...\\
\hspace{3.5mm}evidences: ...\\
\hspace{3.5mm}output: ...\\
\\
\textbf{Your task)}\\
Please read and follow the above instructions and examples step by step\\
query: \{\{QUERY\}\}\\
evidences: \{\{RETRIEVED SUBGRAPHS\}\}\\
\hline
\end{tabular}
\caption{The verbalized subgraph-augemented generation prompt used for KGQA task.}
\label{tab: answer KGQA prompts}
\end{table*}

\begin{table*}[t]
\centering
\small
\begin{tabular}{|>{\arraybackslash}m{14.5cm}|}
\hline
Please verify the statement based on the given evidences from a knowledge graph. \\
\\
\textbf{Notes)}\\
1). If there is any evidence that completely supports the statement, the answer is 'True', otherwise is 'False'.\\
2). For questions like 'A has a wife', if there is any evidence that A has a spouse with any name, the answer is 'True'.\\
3). You should provide a brief reason with several words, then tell that the answer is 'True' or 'False'.\\
\\
\textbf{Examples)}\\
1. query: "Mick Walker (footballer, born 1940) is the leader of 1993–94 Notts County F.C. season."\\
\hspace{3.5mm}evidences: \{\\
\hspace{14mm}"graph [1]": [\\
\hspace{18mm}('Mick Walker (footballer, born 1940)', 'manager', '1993–94 Notts County F.C. season'),\\
\hspace{18mm}('Mick Walker (footballer, born 1940)', 'birthDate', '"1940-11-27"'),\\
\hspace{14mm}],\\
\hspace{14mm}"graph [2]":[\\
\hspace{18mm}('Mick Walker (footballer, born 1940)', 'manager', '1994–95 Notts County F.C. season'),\\
\hspace{18mm}('Mick Walker (footballer, born 1940)', 'birthDate', '"1940-11-27"')\\
\hspace{14mm}]\\
\hspace{3.5mm}\}\\
\hspace{3.5mm}answer: As graphs [1][2] say that Mick Walker is the manager but not the leader, the answer is False.

2. query: ...\\
\hspace{3.5mm}evidences: ...\\
\hspace{3.5mm}output: ...\\
\\
\textbf{Your task)}\\
Please read and follow the above instructions and examples step by step\\
query: \{\{QUERY\}\}\\
evidences: \{\{RETRIEVED SUBGRAPHS\}\}\\
\hline
\end{tabular}
\caption{The verbalized subgraph-augemented generation prompt used in FactKG dataset.}
\label{tab: answer FactKG prompts}
\end{table*}

\section{Implementations for Approaches}
\label{app: implementations}

All programs are implemented with Python.

\subsection{SimGRAG}
Experiments are run with 1 NVIDIA A6000-48G GPU, employing the 4-bit quantized llama3 70B model within the Ollama framework.
We use the Nomic embedding model \cite{nomic}, which generates 768-dim semantic embeddings for nodes and relations.
For retrieving similar nodes (resp. relations), we use HNSW \cite{HNSW} algorithm implemented by Milvus vector database \cite{Milvus}, with maximum degree $M=64$, $efConstruction=512$ and $efSearch=8*k^{(n)}$ (resp. $efSearch=8*k^{(r)}$).

By default, we use $k=3$ and $12$-shot in-context learning throughout all experiments, except for the ablation studies in Section~\ref{sec: ablation}.
For MetaQA dataset, we use $k^{(n)}=k^{(r)}=16$ by default.
For the task of fact verification using FactKG dataset, we use $k^{(n)}=16384$ and $k^{(r)}=512$ by default, except for the grid search that evaluates the retrieval efficiency in Section~\ref{sec: efficacy}.
Moreover, for FactKG dataset, we further utilize the entity type associated with the entity nodes in DBpedia.
Specifically, we construct a mapping that maps a type like ``person'' or ``organization'' to all its entity nodes.
Then, for unknown entities in the pattern graph, such as ``UNKNONWN person 1'', we search for the top-$k^{(t)}$ similar types, then use all nodes with such similar types as the candidate nodes in the retrieval algorithm.
By default, we set $k^{(t)}=16$.

\subsection{Pre-trained LLMs}
For pre-trained LLMs including ChatGPT \cite{ChatGPT} and Llama 3 70B \cite{llama3} without training or augmented knowledge, we also use 12 shots in-context learning for fair comparison.
For Llama 3 70B, experiments are run with 1 NVIDIA A6000-48G GPU, employing the 4-bit quantized model within the Ollama framework.
The license of Llama 3 70B can be found at \url{https://www.llama.com/llama3/license/}.

\subsection{KG-GPT}
For evaluation, we use 1 NVIDIA A6000-48G GPU with the 4-bit quantized Llama3 70B model within the Ollama framework. 
We also use 12-shot in-context learning, and all other parameters are the same as their default setting \cite{kggpt}.

\subsection{KELP}
Experiments were conducted on 1 NVIDIA A6000-48G GPU system. 
Aligned with their settings \cite{KELP}, it involves fine-tuning a 66M-parameter DstilBert model with the AdamW optimizer at a learning rate of $2e-6$ and a batch size of $60$. 
For fairness, we also use 12-shot in-context learning in the prompt.
And we also use Llama 3 70B as the LLM, using the 4-bit quantized model within the Ollama framework.

\subsection{G-Retriever}
Experiments are performed on a system with 6 NVIDIA A6000-48G GPUs. 
The base LLM is the 4-bit quantized llama3 70B with frozen parameters. 
The Graph Transformer served as the GNN, configured with 4 layers, 4 attention heads, and a 1024-dimensional hidden layer. 
During training, we use the AdamW optimizer, a batch size of 4, and 10 epochs, with early stopping after 2 epochs.
All the other parameters are the same with their default settings \cite{G-retriever}.

\subsection{KAPING}
For evaluation, we use 1 NVIDIA A6000-48G GPU with the 4-bit quantized Llama3 70B model within the Ollama framework. 
Aligned with their recommended setting \cite{KAPING}, we retrieve top-10 similar triples using MPNet as the retrieval model.
And their prompt follows a zero-shot approach.

\section{Discussion about Oracle Entities}
\label{app: entity leak}

As discussed in Section~\ref{sec: introduction}, in real applications, users might not always know the precise entity IDs related to their query. 
Thus, an ideal approach should not require users to specify the oracle entities.
However, both KG-GPT \cite{kggpt} and KELP \cite{KELP} expand subgraphs or paths from the user-provided oracle entities, while G-Retriever \cite{G-retriever} restricts the KG to a 2-hop oracle entity neighborhood.
In other words, they need to know which entities are exactly correct before running, and the search space will be constrained in the ground truth area, thereby reducing the problem hardness.

Though all methods will work better with the oracle entities, experimental evaluation in Table~\ref{tab: overall result} shows that even when we allow certain baselines to benefit from using the oracle entities, their performance still underperforms the SimGRAG method that does not require such entities.
In other words, if we do not provide them for such baselines, their performance may degrade further.

Moreover, it is intuitive to use the results of a top-$k$ entity linker as a substitute for the oracle entity in certain baselines. 
However, it would significantly increase the computational complexity and latency, since these methods might need to run the entire pipeline independently for each candidate entity. 
In contrast, the SimGRAG method naturally avoids relying oracle entities without such independent redundant computations.

Furthermore, unlike existing approaches, the internal mechanism of SimGRAG method is designed to better handle and filter out those noisy entities in real-world KGs.
As discussed in Section~\ref{sec: pattern-to-subgraph}, we use the Graph Semantic Distance (GSD) metric, which can effectively incorporate somewhat distant entities or relations that still contribute valuable evidence to the overall subgraph. 
For example, Figure~\ref{fig: semantic similarity} shows that a candidate entity whose semantic distance is ranked $112$ can still be used in the retrieved subgraph.

Generally, the proposed SimGRAG method is closer to the ideal feature through a carefully designed mechanism, while ensuring better performance than the baseline methods.

\section{Detailed Error Analysis}
\label{app: error analysis}

We manually categorize all the encountered errors of the SimGRAG method in our experiments.

The errors occurring during the query-to-pattern alignment step are defined as: LLM fails to follow the given instructions and examples.
For example, for the query ``The lady Anne Monson was born in the Darlington location of the ITL?'' from FactKG dataset, the LLM gives the pattern graph with only one triple ``(`Anne Monson', `born in', `Darlington')'', which is not aligned with the query text.

The error occurred during the subgraph-augmented generation step is defined as that given the correct retrieved subgraph, the LLM fails to provide the final correct response.
For example, for the question ``what films did Lucky McKee star in'' from the MetaQA dataset, correct subgraphs of ``[(`Lucky McKee', `starred\_actors', `Roman')]'' is successfully retrieved, along with the two subgraphs with lower GSD (``[(`Lucky McKee', `directed\_by', `All Cheerleaders Die')]'' and ``[(`Lucky McKee', `directed\_by', `May')]'').
However, the LLM gives the final response of ``According to the evidences, there is no direct connection between Lucky McKee and a film they starred in. The graphs only mention that Lucky McKee directed films (`All Cheerleaders Die' and `May'), but do not provide information about the films they acted in.''

Errors occurring during the pattern-to-subgraph alignment phase are defined as: LLM follows the given instructions and examples to generate a satisfactory pattern graph, but the retrieval algorithm fails to retrieve the ground-truth subgraph for the query.
It is because the ground-truth subgraphs have different structures and thus cannot be successfully aligned with the ground-truth subgraphs.
For example, for the query ``A food is classed as a Dessert and can be served warm (freshly baked) or cold.'',
the LLM-generated pattern graph is ``[(`UNKNOWN food 1', `classed as', `Dessert'), (`UNKNOWN food 1', `served', `"warm"'), (`UNKNOWN food 1', `served', `"cold"')]''.
However, the ground-truth subgraphs have the structure like ``[(`The food name', `classed as', `Dessert'), (`The food name', `served', `"warm (freshly baked) or cold"')]''.

\section{Query Pattern Structures}
\label{app: pattern structures}

Following the previous study \cite{WC2014}, all queries (i.e., the pattern structure) in our experiments can be categorized into the following six types. 
For simplicity, we focus on topological structures and ignore the edge directions.
\begin{itemize}[leftmargin=*]
    \item \textit{1-hop Path:} Find an edge from a known subject $s$ to another entity $e$.
    \item \textit{2-hop Path:} Find a 2-hop path from a known subject $s$ to another known or unknown entity $e$.
    \item \textit{3-hop Path:} Find a 3-hop path from a known subject $s$ to another known or unknown entity $e$.
    \item \textit{2-hop Conjunction:} Find two distinct edges that link two known subjects $s_1$ and $s_2$ with an unknown entity $e$.
    \item \textit{3-hop Conjunction:} Find an edge that links an known subject $s_1$ with an unknown entity $e$, as well as a 2-hop path that connects another known subjects $s_2$ with the same entity $e$.
    \item \textit{3-hop Star:} Find three distinct edges that links to the same known or unknown entities $e$.
\end{itemize}

\section{Parameters for Grid Search}
\label{app: grid search}

We conduct the grid search for evaluating the top-$k$ retrieval algorithm and its optimized one on the FactKG dataset using DBpedia knowledge graph.
Specifically, we randomly sample 100 queries that correctly generate patterns and manually identify the ground truth subgraphs for each query to evaluate retrieval performance using retrieval Hits@1. 
We fix $k=1$ and try all combinations of the other parameters $k^{(n)}\in \{128, 256, 512, 1024, 2048, 4096, 8192, 16384\}$, $k^{(r)}\in \{128, 256, 512\}$, $k^{(t)}\in \{1, 2, 4, 8, 16\}$.
For 100 queries, any program run out of the time limit of 10,000 seconds will be terminated and not reported.
In Figure~\ref{fig: time cost}, the point at retrieval Hits@1=1.0 is achieved by using $k^{(n)}=16384$, $k^{(r)}=512$ and $k^{(t)}=16$.

\section{Experiments on WebQSP Dataset}

We also test SimGRAG on the WebQuestionSP (WebQSP) dataset \cite{WebQSP} using WikiData \cite{wikidata}, which is the most popular and active KG. 
We use the 2015 WikiData dump to align with WebQSP's creation time, which contains 26 million nodes and 57 million edges. 
We use the WebQSP-WD test set \cite{WebQSP-WD}, a corrected version of the original WebQSP dataset for WikiData compatibility. 

Since \cite{WebQSP-WD} mentioned that not all queries are guaranteed answerable with WikiData, we manually verified each query in the test set, excluding those without any supporting evidence in WikiData.
Specifically, each question in the WebQSP-WD test set is associated with a set of topic entities and answer entities, which were mapped from their original Freebase IDs to WikiData IDs. 
Since WebQSP dataset is all about questions within 2-hops \cite{G-retriever}, for each question, we extracted the 2-hop neighborhood of the topic entities, as well as the 2-hop neighborhood of the answer entities in WikiData KG, and union all these edges together to form a single subgraph.
To determine whether a question is supported by WikiData, we compared the question with its corresponding merged subgraph. 
A question was considered unsupported if either of the following conditions held:
\begin{itemize}[leftmargin=*]
    \item None of the topic entities are connected to any answer entities within the subgraph.
    \item All connections between the topic entities and the answer entities are completely irrelevant to the intent of the question. Note that we never require the subgraph to contain a path that directly answers the question. As long as a human could infer the correct answer by reasoning over the whole connected subgraph, we considered the question to be supported.
\end{itemize}

After manually ensuring the quality and reliability of the test set, the proposed SimGRAG method achieves the Hits@1 of 87.7\%.

\end{document}